\documentclass[lettersize,journal]{IEEEtran}
\usepackage{amsmath,amsfonts}
\usepackage{algorithmic}
\usepackage{array}
\usepackage[caption=false,font=normalsize,labelfont=sf,textfont=sf]{subfig}
\usepackage{textcomp}
\usepackage{stfloats}
\usepackage{url}
\usepackage{verbatim}
\usepackage{graphicx}
\usepackage{hyperref}
\usepackage{multirow}
\usepackage{microtype}
\usepackage{booktabs}
\usepackage{xcolor}

\def\BibTeX{{\rm B\kern-.05em{\sc i\kern-.025em b}\kern-.08em
    T\kern-.1667em\lower.7ex\hbox{E}\kern-.125emX}}
\usepackage{balance}
\begin{document}
\title{ELASTIC: Efficient Once For All Iterative Search for Object Detection on Microcontrollers}
\author{Tony Tran, Qin Lin and Bin Hu
\thanks{T. Tran is with the Department of Research Computing, University of Houston, Houston, TX 77004 USA (e-mail: thtran37@CougarNet.UH.EDU).}
\thanks{Q.~Lin, and B. Hu are with the Dept. of Electrical and Computer Engineering, and Dept. of Engineering Technology, University of Houston, Houston, TX 77004 USA (e-mail: qlin21@Central.UH.EDU, bhu11@Central.UH.EDU).}
}

\markboth{TC-2025-10-0949}%IEEE Transactions on Computers,~Vol.~74, No.~10, September~2025
{T. Tran, B. Hu, \MakeLowercase{\textit{(et al.)}:
ELASTIC: Efficient Once For All Iterative Search for Object Detection on Microcontrollers}}

\maketitle

\begin{abstract}
Deploying high-performance object detectors on TinyML platforms poses significant challenges due to tight hardware constraints and the modular complexity of modern detection pipelines. Neural Architecture Search (NAS) offers a path toward automation, but existing methods either restrict optimization to individual modules---sacrificing cross-module synergy---or require global searches that are computationally intractable. We propose \textbf{ELASTIC} (\textbf{\underline{E}}fficient Once for Al\textbf{\underline{L}} Iter\textbf{\underline{A}}tive \textbf{\underline{S}}earch for Objec\textbf{\underline{T}} Detect\textbf{\underline{I}}on on Mi\textbf{\underline{C}}rocontrollers), a unified, hardware-aware NAS framework that alternates optimization across modules (e.g., backbone, neck, and head) in a cyclic fashion. ELASTIC introduces a novel \textit{Population Passthrough} mechanism in evolutionary search that retains high-quality candidates between search stages, yielding faster convergence, \textbf{up to an 8\% final mAP gain}, and eliminates search instability observed without population passthrough. In a controlled comparison, empirical results show ELASTIC achieves \textbf{+4.75\% higher mAP and 2× faster convergence} than progressive NAS strategies on SVHN, and delivers a \textbf{+9.09\% mAP improvement} on PascalVOC given the same search budget. ELASTIC achieves \textbf{72.3\% mAP} on PascalVOC, outperforming \textbf{MCUNET by 20.9\%} and \textbf{TinyissimoYOLO by 16.3\%}. When deployed on MAX78000/MAX78002 microcontrollers, ELASTIC-derived models outperform Analog Devices’ TinySSD baselines, reducing energy by up to \textbf{71.6\,\%}, lowering latency by up to \textbf{2.4$\times$}, and improving mAP by up to \textbf{6.99} percentage points across multiple datasets. {The experimental videos and codes are available on the
\href{https://nail-uh.github.io/elastic.github.io/}{project website}\footnote{Project Website: \url{https://nail-uh.github.io/elastic.github.io/}}}.
\end{abstract}

\begin{IEEEkeywords}
Neural Architecture Search, TinyML, Object Detection, Population Passthrough, Iterative Evolutionary Architecture Search
\end{IEEEkeywords}

\section{Introduction}

Object detection is a fundamental task in computer vision with real-world applications in mobile robotics, autonomous vehicles, smart surveillance, and embedded systems. While modern detectors achieve remarkable accuracy and speed on powerful hardware, their deployment on \emph{TinyML platforms}—microcontrollers and edge devices with sub-MB memory, ultra-low power, and limited compute budgets—remains challenging~\cite{lin_mcunet_2020, liang_mcuformer_2023, moosmann_tinyissimoyolo_2023, moss_ultra-low_2022, yan_lighttrack_2021}.

% TinyML enables always-on inference with stringent energy and latency constraints~\cite{wong_tiny_2018, moosmann_flexible_2023}. 
% However, unlike image classification, object detection involves multi-stage pipelines—typically composed of a \emph{backbone}, \emph{neck}~\cite{ghiasi_nas-fpn_2019}, and \emph{head}—each contributing uniquely to accuracy, latency, and memory footprint. These interdependent components enlarge the design space exponentially, complicating deployment on microcontrollers, where model size, SRAM usage, and inference cycles must be tightly controlled~\cite{wang_bed_2022, moosmann_tinyissimoyolo_2023, burrello_dory_2021}.

Unlike image classification, object detection relies on a multi-stage pipeline—\emph{backbone}, \emph{neck}~\cite{ghiasi_nas-fpn_2019}, and \emph{head}—whose design choices are strongly interdependent: modifying one module changes the feature distributions and optimal configurations of the others. This cross-module coupling makes the detector design space highly combinatorial, which is particularly challenging on microcontrollers where flash/SRAM budgets and inference cycles must be tightly controlled~\cite{wang_bed_2022, moosmann_tinyissimoyolo_2023, burrello_dory_2021}. Recent hardware-aware NAS efforts for \emph{pure MCU} targets have largely focused on \emph{image classification}~\cite{Garavagno2024Affordable, Garavagno2025Searching}, where the optimization is over single-stream networks with a single prediction head and classification objectives; in contrast, MCU detection requires jointly reasoning over the coupled backbone--neck--head pipeline and mAP-driven evaluation.

Recent TinyML-oriented detection research improves efficiency via specialized architectures and deployment-aware design. Hardware-conscious model and compilation/runtime co-design enable practical MCU execution~\cite{lin_mcunet_2020, burrello_dory_2021, wang_bed_2022}, while lightweight detectors and variants (including transformer-inspired designs) explore accuracy--efficiency trade-offs under tight energy/latency constraints~\cite{moosmann_tinyissimoyolo_2023, moss_ultra-low_2022, yan_lighttrack_2021, liang_mcuformer_2023}. Despite these advances, selecting an optimal detector for a specific MCU remains non-trivial because accuracy, flash/SRAM footprint, and inference cycles are tightly coupled, and manual design or ad-hoc scaling rarely navigates these constraints systematically at the full detector level.

\emph{Neural Architecture Search} (NAS) offers a compelling approach to automate architecture design under tight resource constraints. Constraint-aware NAS has advanced edge classification~\cite{cai_once-for-all_2020, lin_mcunet_2020, lv_-situ_2025}, but object detection adds new layers of complexity. Prior detection NAS methods focus on individual modules (e.g., backbones~\cite{chen_detnas_2019}, necks~\cite{ghiasi_nas-fpn_2019}, or heads~\cite{wang_nas-fcos_2020}), often ignoring cross-module dependencies. Full-pipeline NAS is possible~\cite{guo_hit-detector_2020}, but scales poorly due to the massive combinatorial space—e.g., over $10^{28}$ configurations in realistic search settings~\cite{wang_nas-fcos_2020, sakuma_detofa_2023}. These observations expose a key gap: MCU-targeted detection NAS must preserve cross-module co-adaptation (backbone--neck--head interactions) without incurring the prohibitive cost of full end-to-end search, while enforcing hard resource constraints so discovered models are directly deployable. 

%ELASTIC addresses this by cyclically alternating module optimization, enabling repeated co-adaptation under explicit memory/compute budgets.

% \textcolor{blue}{These observations expose a key gap: an effective MCU-targeted detection NAS method must capture \emph{cross-module} interactions (backbone--neck--head co-adaptation), yet avoid the prohibitive cost of searching the full end-to-end space. Module-wise searches are efficient but can miss globally optimal designs because the best choice for one module depends on the others; conversely, full-pipeline search becomes intractable as the joint configuration space explodes~\cite{guo_hit-detector_2020, wang_nas-fcos_2020, sakuma_detofa_2023}. What is needed is a search strategy that (i) preserves interdependencies via repeated co-adaptation of modules, (ii) reduces complexity by decomposing optimization into manageable steps, and (iii) enforces explicit resource constraints throughout so that discovered architectures are deployable on real microcontrollers.}

To address these limitations, we introduce \textbf{ELASTIC} (\textbf{\underline{E}}fficient Once for Al\textbf{\underline{L}} Iter\textbf{\underline{A}}tive \textbf{\underline{S}}earch for Objec\textbf{\underline{T}} Detect\textbf{\underline{I}}on on Mi\textbf{\underline{C}}rocontrollers), a novel once-for-all NAS method that performs \emph{constrained iterative search}. ELASTIC alternates between optimizing one module while fixing others in a cyclic fashion, leveraging previous iteration feedback to refine architecture choices. This strategy preserves cross-module interdependencies while dramatically reducing the effective search space. Further, our approach enforces explicit resource constraints—such as model size—throughout the search process, ensuring deployability on MCUs.

Our key contributions are summarized as follows:
\begin{itemize}
    \item \textbf{Unified Iterative NAS}: We propose a cyclic module-wise NAS framework that alternates optimization across the backbone, neck, and head to preserve cross-module co-adaptation while reducing search cost by restricting the active search dimension at each iteration. In our early-stage PascalVOC search, ELASTIC achieves up to \textbf{+4.75\% mAP} over the progressive baseline while on SVHN, ELASTIC reduces search cost until convergence from \textbf{30.8 to 12.5 GPU hours}.
    \item \textbf{Population Passthrough Mechanism:} We introduce \emph{Population Passthrough}, an elite-carryover strategy that stabilizes the \emph{iterative} module alternation by retaining top-performing candidates when switching the active search module, rather than reinitializing the evolutionary population. This directly mitigates the abrupt performance drops observed under naive module switching and enables consistent convergence across iterations. On PascalVOC, enabling passthrough improves the final mAP from \textbf{22.1\%} (iterative search \emph{without} passthrough) to \textbf{30.83\%} (with passthrough), demonstrating that the mechanism is a necessary component for making cyclic multi-module optimization effective in practice.
    \item \textbf{Enhanced Search Space:} ELASTIC enhances the search space over time. On SVHN, the mean mAP of randomly sampled architectures improves from \textbf{44.61\% (joint)} to \textbf{70.68\%} after 5 iterations, while variance reduces by \textbf{89.7\%}. On PascalVOC, the mean mAP improves from \textbf{4.99\%} to \textbf{28.32\%}, with a \textbf{95.3\% variance reduction}.
    \item  \textbf{Once-for-All Deployment on Resource-Constrained Devices}: We validate ELASTIC in designing versatile models tailored to various resource constraints, demonstrating their inference efficiency on the MAX78000/02 and STM32F746 platform. Our ELASTIC model significantly outperforms the baseline \cite{ai85tinierssd} on the MAX78000 platform, achieving a \textbf{45.4\% reduction} in energy consumption, a \textbf{29.3\% decrease} in latency, and a \textbf{4.5\% improvement} in mAP, demonstrating enhanced computational efficiency and accuracy.
\end{itemize}

\section{Related Work}

\textbf{NAS for Object Detection.}
Object detection architectures are modular, comprising a backbone, neck, and head. NAS methods have targeted these components individually or jointly: DetNAS~\cite{chen_detnas_2019} searches backbone architectures via a one-shot supernet; NAS-FPN~\cite{ghiasi_nas-fpn_2019} focuses on FPN design; NAS-FCOS~\cite{wang_nas-fcos_2020} extends search to the FPN and head using reinforcement learning; MobileDets~\cite{xiong_mobiledets_2021} introduces latency-aware backbone optimization for mobile accelerators; and Hit-Detector~\cite{guo_hit-detector_2020} jointly searches all modules but incurs high computational cost. While effective on larger systems, these methods are not designed for deployment on resource-constrained microcontrollers.

\textbf{Progressive vs. Global Search.}
Global NAS jointly optimizes all modules but is intractable at TinyML scales due to the combinatorial size of the joint search space~\cite{guo_hit-detector_2020, wang_nas-fcos_2020}. Progressive NAS reduces complexity by optimizing modules sequentially—typically starting with the backbone or neck—while keeping previously optimized components fixed~\cite{wang_nas-fcos_2020, rana_nas-od_2024}. However, early architectural choices (e.g., output resolution, feature dimensionality, or stage depths) impose rigid interface constraints that limit the feasible design space for subsequent modules, often leading to suboptimal detection accuracy or violating tight memory and compute budgets in later stages~\cite{ghiasi_nas-fpn_2019}. Our approach addresses this by iteratively alternating module optimization, enabling cross-stage adaptation while preserving search efficiency.

\textbf{OFA and Iterative NAS.}
Once-for-all (OFA) methods train a supernet capable of representing many sub-networks, enabling fast specialization to various hardware or latency constraints without retraining~\cite{cai_once-for-all_2020}. DetOFA~\cite{sakuma_detofa_2023} adapts this strategy to object detection using constraint-aware path pruning. These methods provide a foundation for efficient, hardware-aware NAS; our work builds on this by incorporating modular search dynamics specific to detection pipelines.

\begin{figure*}[t]
    \centering
    \includegraphics[width=0.75\linewidth]{imgs/method.png}
    \caption{\textbf{Overview of ELASTIC}: Our method begins with a pretrained supernet and performs iterative neural architecture search by alternating optimization between the backbone and head. The Population Passthrough mechanism ensures continuity by retaining top-performing candidates across module alternations.}
    \label{fig:method}
\end{figure*}

\section{Methodology}
\label{sec:methodology}

This section introduces \textbf{ELASTIC}, a constrained iterative NAS method for \emph{TinyML object detection} that explicitly treats the detector as a \emph{multi-module pipeline} (backbone--neck--head). The key distinction from prior NAS strategies is that ELASTIC performs \emph{cyclic module-wise optimization}: at each iteration, we optimize one module while keeping the others fixed, and we revisit modules multiple times to enable \emph{cross-module co-adaptation} without incurring the prohibitive cost of full end-to-end search. However, naively switching the optimized module can destabilize the search and cause abrupt drops in validation performance. To address this, we propose \textbf{Population Passthrough}, which retains top-performing architecture candidates at the end of each module stage and injects them into the next stage, stabilizing convergence across iterations under explicit resource constraints.

\subsection{Problem Formulation}
\label{formulation}

Let an object detection subnet be represented as $\mathcal{N}(f, w)$ extracted from a supernet $\mathcal{N}$ with architecture $f$ and weights $w$. The search space $\mathcal{A}$ contains all valid architectures such that $f \in \mathcal{A}$. The objective of NAS is to find the optimal architecture $f^*$ that minimizes the validation loss $\mathcal{L}_{val}$ when evaluated with shared weights $W^*$:

\begin{equation}
    \min_{f \in \mathcal{A}} \mathcal{L}_{\text{val}}(\mathcal{N}(f, W^*(f)))
    \label{eq1}
\end{equation}
\begin{equation}
    \text{s.t.}\quad
    W^* = \arg\min_{w} \sum_{f \in \mathcal{A}} \mathcal{L}_{\text{train}}(\mathcal{N}(f, w))
    \label{eq2}
\end{equation}
\begin{equation}
    |W^*(f)| \leq \tau
    \label{eq3}
\end{equation}

Here, $\tau$ is the hardware-constrained model size. Inspired by Once-for-All NAS~\cite{cai_once-for-all_2020}, a weight-shared supernet reduces training cost. Since $\mathcal{A}$ grows combinatorially for joint optimization (e.g., $f=(b,h)$), optimization alternates between backbone $b$ and head $h$. To extend this formulation to neck modules, the neck design choices can be integrated into the head search space, enabling joint optimization of both neck and head components within a single search stage.

At iteration $t$, we first optimize the backbone while keeping $h^{(t)}$ fixed:

\begin{equation}
    b^{(t+1)} = \arg\min_{b} \mathcal{L}_{\text{val}}(\mathcal{N}((b, h^{(t)}), w^*))
\end{equation}
\begin{equation}
    \text{s.t.}\quad
    w^* = \arg\min_{w} \sum_{f \in \mathcal{A}} \mathcal{L}_{\text{train}}(\mathcal{N}(f, w))
\end{equation}
\begin{equation}
    |w_b| \leq \tau_b
    \label{eq4}
\end{equation}

Then, we fix $b^{(t+1)}$ and update the head:

\begin{equation}
    h^{(t+1)} = \arg\min_{h} \mathcal{L}_{\text{val}}(\mathcal{N}((b^{(t+1)}, h), w^*))
\end{equation}
\begin{equation}
    \text{s.t.}\quad
    w^* = \arg\min_{w} \sum_{f \in \mathcal{A}} \mathcal{L}_{\text{train}}(\mathcal{N}(f, w))
\end{equation}
\begin{equation}
    |w_h| \leq \tau_h
    \label{eq5}
\end{equation}

We enforce the constraint $\tau_b + \tau_h \leq \tau$, where $\tau$ is the global memory budget. This formulation ensures that each module’s optimization phase respects deployment constraints.

\subsection{Iterative Evolutionary Architecture Search}
\label{iterativeea}

At the core of ELASTIC is an \textit{Iterative Evolutionary Architecture Search} that alternates optimization between the backbone and the head over successive cycles. As illustrated in Fig.~\ref{fig:method}, the framework does not search the full detector space in a single pass. Instead, it fixes one module while optimizing the other, then switches roles and repeats this process iteratively. This design preserves search tractability for TinyML deployment while allowing the two modules to progressively co-adapt across iterations.

Within each stage shown in Fig.~\ref{fig:method}, the evolutionary search follows a standard population-based procedure, but is applied only to the currently active module. The population is initialized using two sources: high-performing candidates carried over from the module's previous visit through \textit{Population Passthrough}, and newly sampled candidates that maintain exploration. New architectures are then generated through crossover and mutation under the corresponding resource constraints (e.g., $\tau_b$ for the backbone and $\tau_h$ for the head). Each candidate is evaluated inside the weight-shared supernet using validation mAP, where the complementary module remains fixed. The best-performing candidates are retained to guide the next generation and to provide a strong starting point when the search returns to that module in a later cycle.

As depicted in Fig.~\ref{fig:method}, once the search for the current module is completed, ELASTIC switches to the other module and repeats the same procedure. By alternating in this back-and-forth manner, the framework incrementally refines both modules rather than optimizing them only once. This repeated revisiting is important because improving one module changes the context in which the other operates. Therefore, the iterative structure of ELASTIC enables the backbone and head to adapt to each other over time while remaining within deployment constraints.

\subsection{Population Passthrough}
\label{passthrough}

% A central challenge in iterative modular NAS is maintaining optimization continuity across alternating search phases. When a search process shifts focus from one module (e.g., backbone) to another (e.g., head), conventional evolutionary algorithms often discard the entire population from the previous cycle. This reinitialization disrupts search momentum, leading to slower convergence and redundant exploration. 

\begin{figure*}[t]
  \centering
  \includegraphics[width=0.75\linewidth]{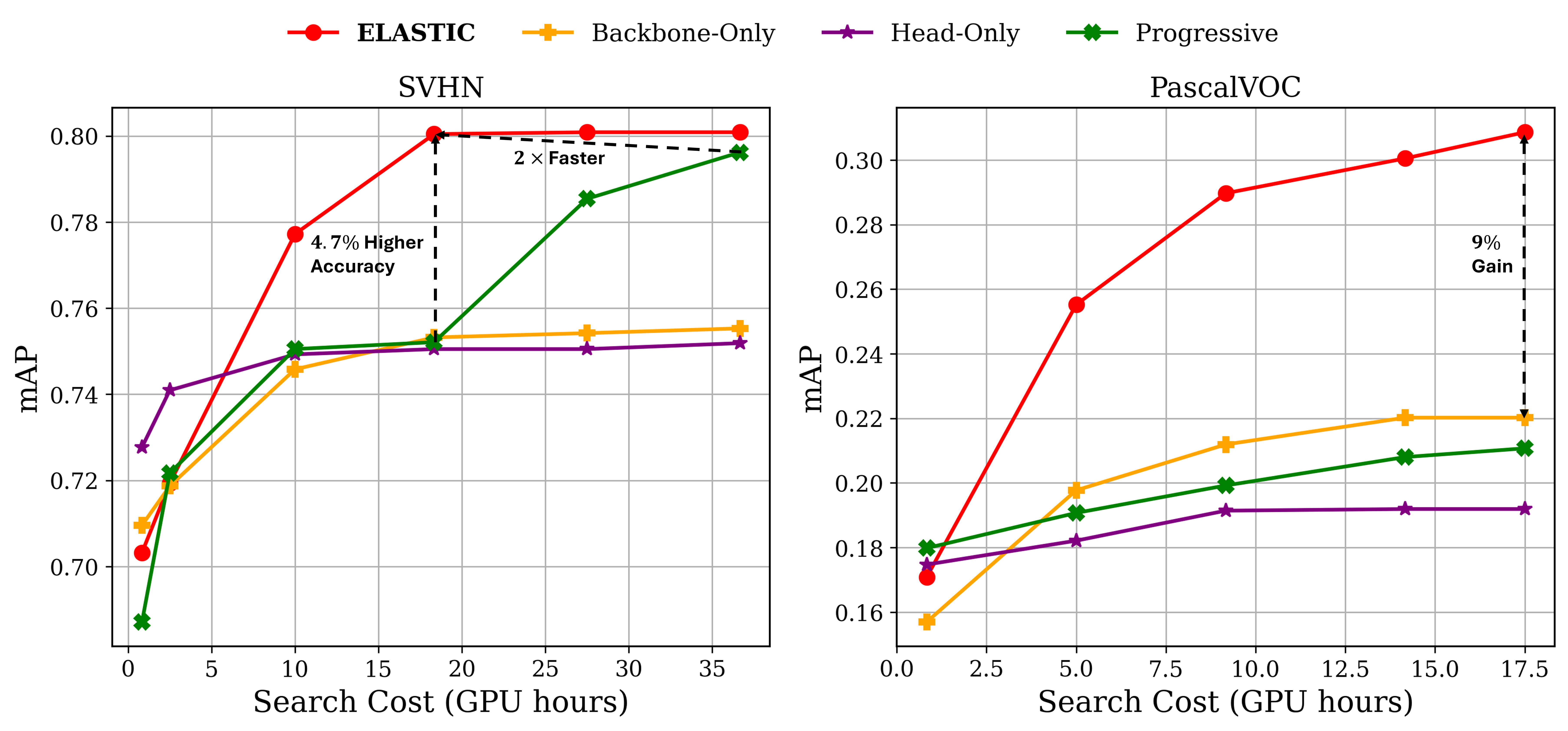}
  \caption{
    \textbf{Comparison of Neural Architecture Search (NAS) strategies on the SVHN (left) and PascalVOC (right) subset.}
        On the SVHN dataset, \textsc{ELASTIC} achieves \textbf{4.7\% higher mAP} and \textbf{2$\times$ faster} than Progressive. On PascalVOC, \textsc{ELASTIC} achieves a \textbf{9\% absolute mAP gain} over all other approaches in 17.5 GPU hours.}
  \label{fig:history}
\end{figure*}

\begin{table*}[t]
    \centering
    \small
    \caption{\textbf{Quantitative comparison of search strategies on the SVHN and PascalVOC subset.} ELASTIC consistently achieves the highest mAP on both datasets and reduces GPU hours by 59\% on SVHN.}
    \begin{tabular}{llcccccc}
        \toprule
        Dataset & Method 
        & mAP & $\uparrow$mAP & Cost & MACs & Params & Latency \\
        \midrule
        \multirow{4}{*}{SVHN}
        & Backbone-Only~\cite{chen_detnas_2019, chen2025hionas, sakuma2023detofa} & 75.53\% & +0.18\% & 10.8\,hrs & 137.6\,M & 0.73\,M & 2.62\,ms \\
        & Neck/Head-Only~\cite{ghiasi_nas-fpn_2019, kang2025research, xue2025easfp}   & 75.35\% & +0.00\% & 4.7\,hrs & 475.3\,M & 1.01\,M & 2.69\,ms \\
        & Progressive~\cite{wang_nas-fcos_2020} & 79.62\% & +4.27\% & 30.8\,hrs & 78.4\,M & 0.52\,M & 2.36\,ms \\
        & \textbf{ELASTIC (OURS)}                      & \textbf{80.09\%} & \textbf{+4.74\%} & \textbf{12.5\,hrs} & 105.8\,M & 0.61\,M & \textbf{2.34\,ms} \\
        \midrule
        \multirow{4}{*}{PascalVOC}
        & Backbone-Only~\cite{chen_detnas_2019} & 22.02\% & +2.80\% & 9.6\,hrs & 436.1\,M & 0.58\,M & 2.11\,ms \\
        & Neck/Head-Only~\cite{ghiasi_nas-fpn_2019, kang2025research, xue2025easfp}   & 19.22\% & +0.00\% & 6.0\,hrs & 240.0\,M & 0.36\,M & 2.48\,ms \\
        & Progressive~\cite{wang_nas-fcos_2020} & 21.74\% & +2.52\% & 14.8\,hrs & 540.0\,M & 0.41\,M & 2.66\,ms \\
        & \textbf{ELASTIC (OURS)}                      & \textbf{30.83\%} & \textbf{+11.61\%} & \textbf{14.65\,hrs} & 642.8\,M & 0.57\,M & \textbf{2.00\,ms} \\
        \bottomrule
    \end{tabular}
    \label{tab:main_results}
\end{table*}

\begin{table*}[t]
    \centering
    \small
    \caption{\textbf{Comparison of ELASTIC with TinyissimoYOLO~\cite{moosmann_flexible_2023} and MCUNET~\cite{lin_mcunet_2020, lin_mcunetv2_2024} on the full PascalVOC dataset, considering all object classes and counts.} ELASTIC achieves a 20.9\% mAP boost over MCUNET~\cite{lin_mcunet_2020}, 4.0\% over MCUNetV2~\cite{lin_mcunetv2_2024}, and a 16.3\% over TinyissimoYOLO’s~\cite{moosmann_flexible_2023} best performing model, while discovering a model with significantly fewer MACs, enabling faster inference.} 
    \begin{tabular}{lccccc}
        \toprule
        Method & MACs & $\downarrow$MACs & Params & VOC mAP & $\uparrow$mAP \\
        \midrule
        TY: 20-3-88 \cite{moosmann_flexible_2023} & 32M & 90.7\% & 0.58M & 53\% & +30\% \\
        TY: 20-7-88 \cite{moosmann_flexible_2023} & 44M & 87.2\% & 0.58M & 47\% & +24\% \\
        TY: 20-3-112 \cite{moosmann_flexible_2023} & 54M & 84.3\% & 0.89M & 56\% & +33\% \\
        TY: 20-7-112 \cite{moosmann_flexible_2023} & 70M & 79.6\% & 0.91M & 53\% & +30\% \\
        TY: 20-3-224 \cite{moosmann_flexible_2023} & 218M & 36.4\% & 3.34M & 23\% & +0\% \\
        MCUNet \cite{lin_mcunet_2020} & 168M & 51.0\% & 1.2M & 51.4\% & +28.4\% \\
        MCUNetV2-M4 \cite{lin_mcunetv2_2024} & 172M & 49.9\% & 0.47M & 64.6\% & +41.6\% \\
        MCUNetV2-H7 \cite{lin_mcunetv2_2024} & 343M & 0\% & 0.67M & 68.3\% & +45.3\% \\
        \textbf{ELASTIC (OURS)} & 
        86M & 
        74.9\% & 
        1.36M & 
        \textbf{72.3\%} & 
        \textbf{+49.3\%} \\
        \bottomrule
    \end{tabular}
    \label{tab:mcunet}
\end{table*}

A key practical challenge in ELASTIC’s \emph{outer-loop} cyclic module-wise optimization is maintaining optimization continuity when switching the \emph{active} module (e.g., backbone $\rightarrow$ neck/head $\rightarrow$ backbone). Although the \emph{inner-loop} evolutionary optimizer is standard, naive alternation effectively \emph{cold-starts} each stage by discarding the module-specialized population, forcing repeated re-discovery of candidates compatible with the fixed context.

To make alternation workable, we introduce \textbf{Population Passthrough}, a lightweight memory mechanism that transfers elite architectural ``knowledge'' across module switches. For each module $m \in \{b, h\}$ (backbone vs.\ neck/head), we maintain a \emph{passthrough buffer} $\mathcal{M}_m$ containing the top-performing candidates (ranked by validation mAP under the current resource constraints) from the most recent optimization of $m$. When ELASTIC revisits $m$, we initialize the population by mixing a fixed fraction of elites from $\mathcal{M}_m$ (inheritance) with newly sampled/mutated candidates (diversity), balancing exploitation and exploration.

This mechanism is \emph{new in our setting} because it targets \emph{alternating modular} NAS: unlike standard elitism/aging schemes that assume a single, fixed search space and continuous run, Population Passthrough is applied \emph{at module boundaries} to stabilize the outer-loop alternation and avoid repeated re-discovery.

\subsection{Resource Allocation and Search Budget}

To ensure deployability on resource-constrained hardware, we explicitly partition the overall memory budget $\tau$ between architectural modules. Let $\tau_b$ and $\tau_h$ denote the memory allocations for the backbone and head, respectively, such that $\tau_b + \tau_h \leq \tau$. This modular budget decomposition is critical for embedded systems, where hardware constraints—such as on-chip SRAM, flash storage, and runtime latency—are rigid and non-negotiable~\cite{moss_ultra-low_2022}. By enforcing per-module constraints during search, ELASTIC ensures that all discovered subnetworks are compatible with the deployment requirements of target TinyML platforms.

In addition, ELASTIC is designed to support \textit{search interruption}, allowing the iterative process to be paused at any stage and a valid, high-quality subnetwork to be extracted. This property is particularly beneficial in real-world TinyML development workflows, where compute resources are limited and iterative refinement may need to occur incrementally~\cite{liang_mcuformer_2023}. 

\section{Experiments}
\label{sec:experiment}

% Our experimental evaluation of ELASTIC is organized into five parts: Section~\ref{sec:setup} describes the setup and training of modular supernets; Section~\ref{sec:comparison} presents a comparative study of search performance against baseline methods; Section~\ref{sec:passthrough} investigates the Population Passthrough technique; Section~\ref{sec:refinement} analyzes how the search space improves over iterations; and Section~\ref{sec:deployment} demonstrates deployment results on resource-limited devices.

\begin{figure}[t]
    \centering
    \includegraphics[width=0.8\linewidth]{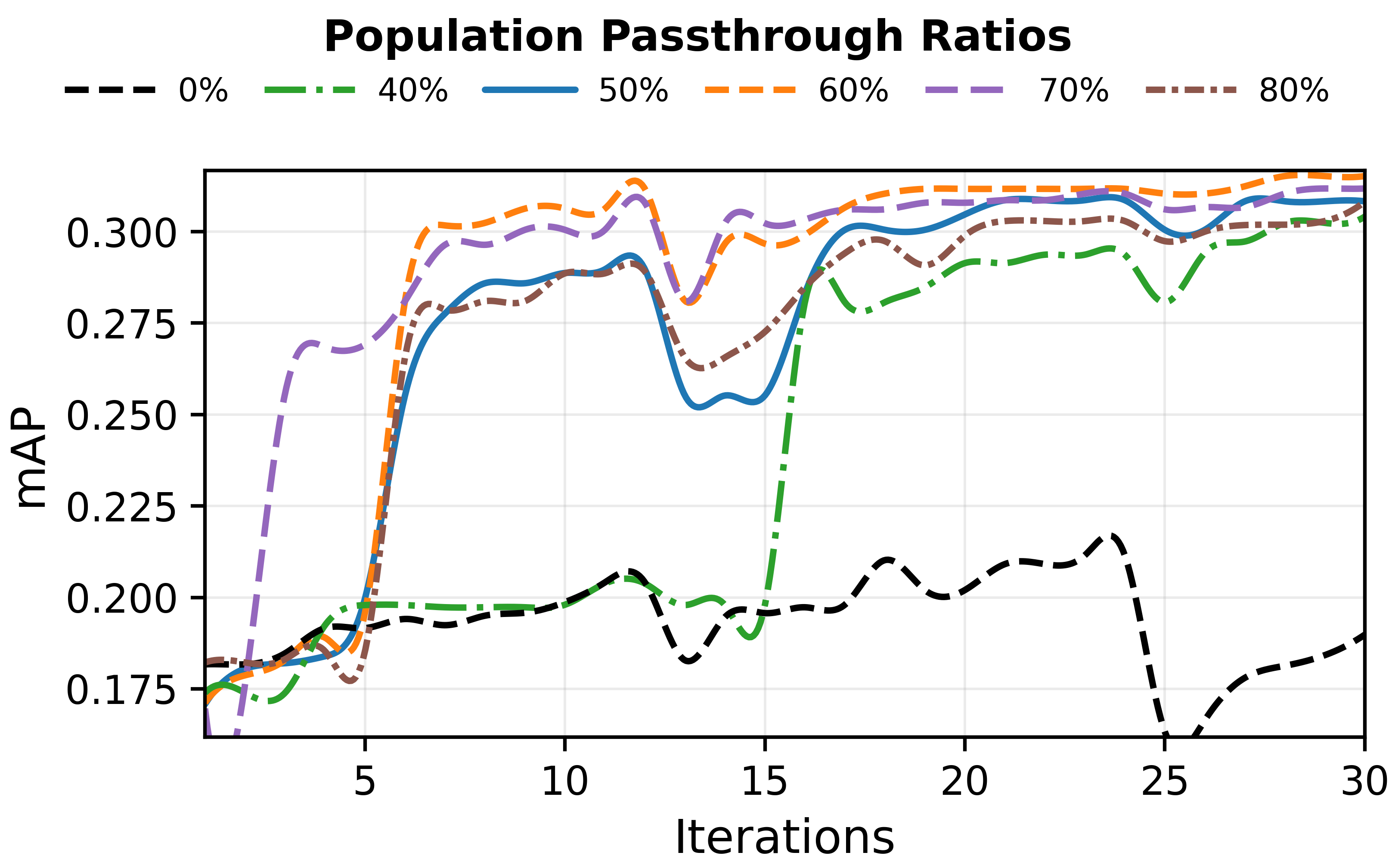}
    \caption{\textbf{Effect of Population Passthrough on Search Stability.} Eliminating passthrough (0\%) leads to unstable search dynamics, with mAP repeatedly dropping and stalling at 22.1\%. Introducing a moderate passthrough rate (50--80\%) stabilizes training and improves convergence, achieving over 30\% mAP (+8\%).}
    \label{fig:passthrough}
\end{figure}

\begin{figure*}[t]
    \centering
    \includegraphics[width=0.75\textwidth]{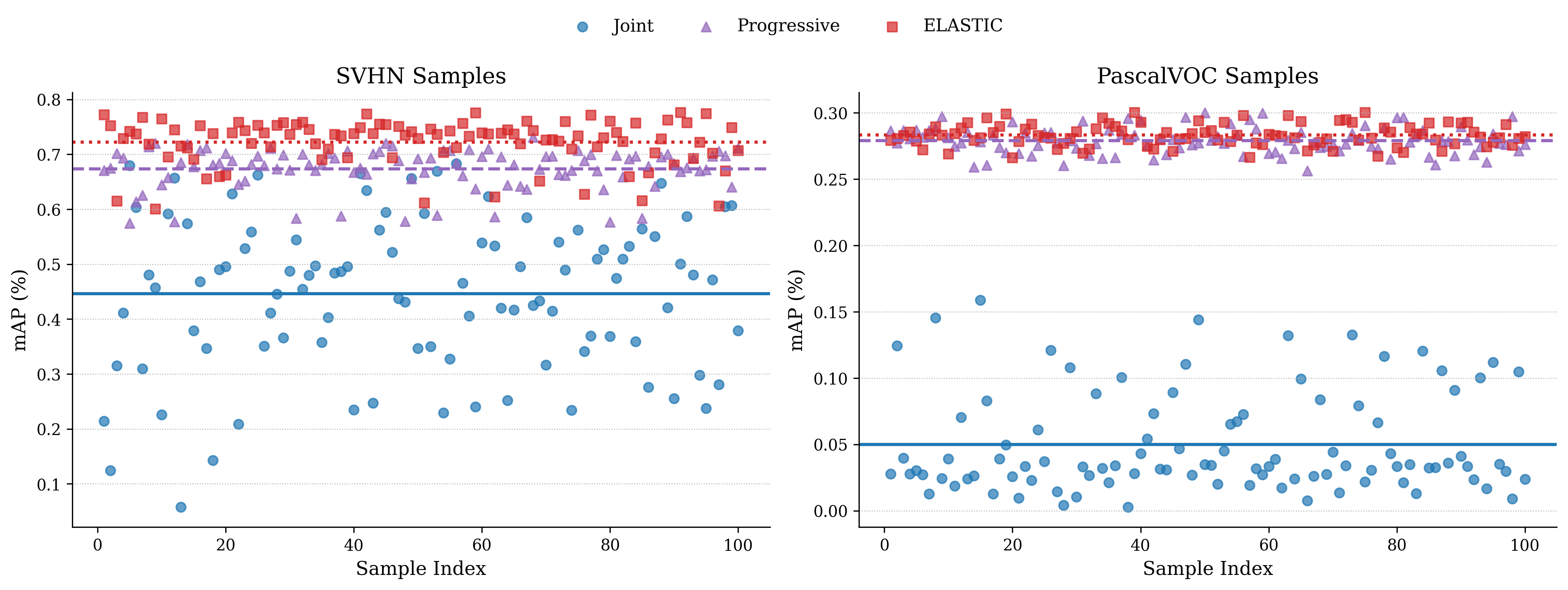} % height=4cm
    \caption{\textbf{Search space refinement through ELASTIC iteration.} Distributions of 100 randomly sampled architectures from three head search spaces—joint NAS, progressive head-only, and ELASTIC-refined—on SVHN (left) and PascalVOC (right). ELASTIC produces architectures with a mean mAP improvement of +4.87\% on SVHN and +0.43\% on PascalVOC over progressive search. Compared to global joint search, ELASTIC improves the mAP by 26.07\% and 23.33\% on SVHN and PascalVOC, respectively.}
    \label{fig:sampling}
\end{figure*}

\begin{figure*}[t]
  \centering
  \includegraphics[width=0.75\textwidth]{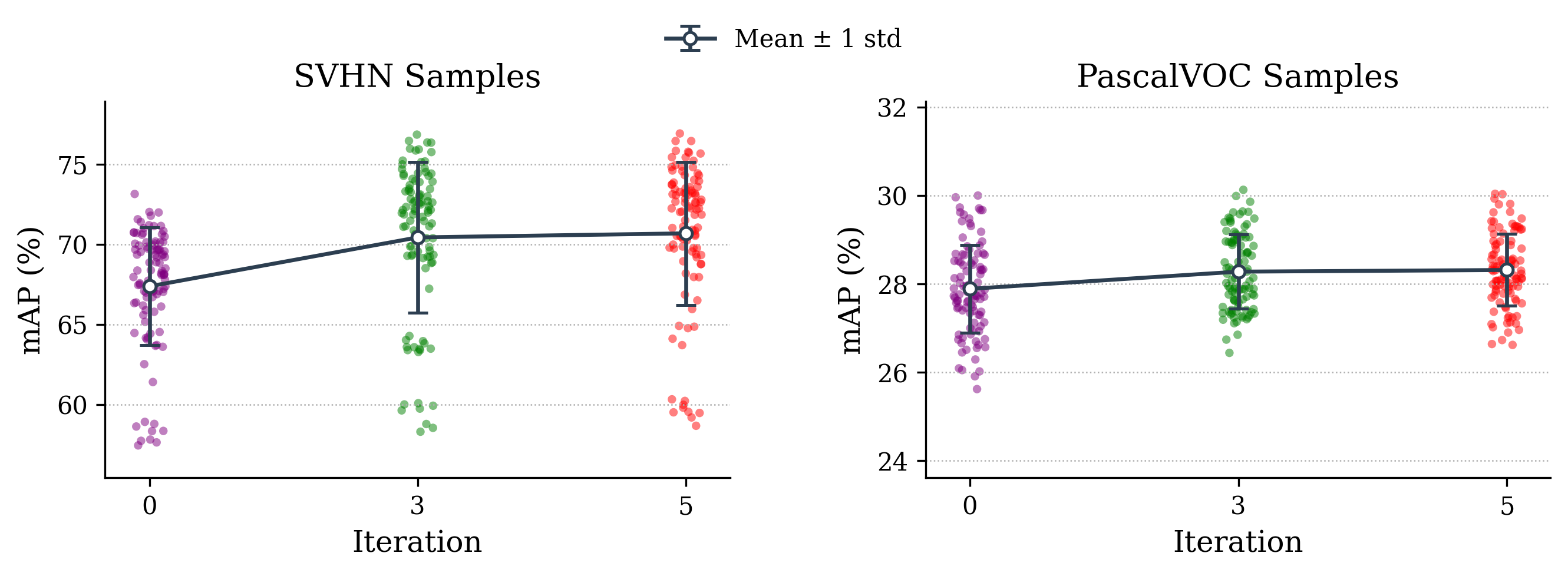}
  \caption{
    \textbf{Search Space Evolution Over Iterations.} 
    Mean accuracy and distribution of sampled architectures at iterations 0, 3, and 5. On SVHN, mean mAP improves by 3.3\% and variance drops by 89.7\%; on PascalVOC, mean mAP rises from 27.89\% to 28.32\% and variance decreases by 33\%.}
  \label{fig:errorplot}
\end{figure*}

\begin{table*}[t]
    \centering
    \small
    \setlength{\tabcolsep}{4pt}
    \renewcommand{\arraystretch}{1.1}
    \caption{\textbf{Mean mAP and variance across search spaces.} Superscripts denote iterative search iterations. Higher mean mAP is better; lower variance indicates more consistent architectures.}
    \begin{tabular}{llcccc}
        \toprule
        Dataset & Search Space & Mean mAP & $\uparrow$mAP & Std. Dev. & Variance \\
        \midrule
        \multirow{4}{*}{SVHN} 
        & Joint Search~\cite{guo_hit-detector_2020} & 44.61\% & 0\% & 1.39 & $1.93$ \\
        & Neck/Head Search~\cite{ghiasi_nas-fpn_2019, kang2025research, xue2025easfp}     & 67.38\% & +22.77\% & \textbf{0.37} & \textbf{0.136} \\
        & \textbf{ELASTIC}$^{(3)}$                       & 70.42\% & +25.81\% & 0.47 & 0.221 \\
        & \textbf{ELASTIC}$^{(5)}$                       & \textbf{70.68\%} & \textbf{+26.07\%} & 0.45 & 0.199 \\
        \midrule
        \multirow{4}{*}{PascalVOC}
        & Joint Search~\cite{guo_hit-detector_2020} & 4.99\%  & 0\% & 0.04 & $1.40 \times 10^{-3}$ \\
        & Neck/Head Search~\cite{ghiasi_nas-fpn_2019, kang2025research, xue2025easfp}     & 27.89\% & +22.90\% & \textbf{0.01} & \textbf{$9.86 \times 10^{-5}$} \\
        & \textbf{ELASTIC}$^{(3)}$                 & 28.28\% & +23.29\% &\textbf{0.01} & \textbf{$7.04 \times 10^{-5}$} \\
        & \textbf{ELASTIC}$^{(5)}$                 & \textbf{28.32\%} & \textbf{+23.33\%} & \textbf{0.01} & $\mathbf{6.60 \times 10^{-5}}$ \\
        \bottomrule
    \end{tabular}
    \label{tab:mean_std_variance}
\end{table*}

\subsection{Experimental Setup}
\label{sec:setup}

We construct five OFA-based supernets tailored for diverse TinyML-relevant datasets: (i) \emph{SVHN}, (ii) \emph{VGGFace2}, (iii) a five-class subset of \emph{PascalVOC}, (iv) the full \emph{PascalVOC} dataset, and (v) \emph{TrashNet}. These datasets are selected to cover a range of TinyML application domains, from lightweight digit recognition to face identification and general-purpose object detection, while avoiding the prohibitive memory and compute demands of large-scale datasets such as COCO.

The \emph{SVHN}, \emph{VGGFace2}, and \emph{PascalVOC subset} supernets adopt a one-stage SSD-style architecture, while the \emph{full PascalVOC} and \emph{TrashNet} supernets adopt a MobileNet--YOLO style design that combines a lightweight MobileNet backbone with YOLO-style detection heads. For the SSD-style supernets, the search space varies (i) layer widths, (ii) kernel sizes ($1\times1$, $3\times3$), and (iii) depth, i.e., the number of layers allocated at each feature map. For both the MobileNet--YOLO supernets, the search space varies (i) expansion ratios, (ii) kernel sizes ($1\times1$, $3\times3$), and (iii) depth, i.e., the number of blocks allocated at each feature map.

All supernets are trained with a batch size of 64, learning rate of 0.001, momentum of 0.9, and weight decay of $5\times10^{-4}$. Training follows the progressive shrinking strategy~\cite{cai_once-for-all_2020}, with 200 epochs per level transition. Experiments are conducted on a single NVIDIA RTX 4090 GPU.

We evaluate deployment across three MCU platforms. The MAX78000/02 are accelerator-equipped microcontrollers with a dedicated CNN engine and separate on-chip memory for weights and activations. MAX78000 provides 442\,kB weight memory and 524\,kB data memory for the earliest streaming layers, while subsequent layers must fit within a single 32\,kB memory instance. MAX78002 provides 2.4\,MB weight memory and 1.3\,MB data memory for the earliest streaming layers, while subsequent layers must fit within a single 80\,kB memory instance. In contrast, the STM32F746 is a general-purpose microcontroller without a CNN accelerator, with 320\,kB SRAM and 1\,MB flash, where weights are stored in flash and activations are stored in SRAM. In this way, MAX78000/MAX78002 represent accelerator-equipped TinyML targets, while STM32F746 provides a non-accelerator MCU baseline for comparison.

Evolutionary search is performed with a population size of 100, mutation probability of 0.2, mutation ratio of 0.5, and a parent selection ratio of 0.25. Each search is budgeted for 55 generations on SVHN and VGGFace2, and 30 generations on the PascalVOC subset, full PascalVOC, and TrashNet datasets, respectively. Convergence is defined as the point at which the best candidate architecture’s mAP is within 1\% of the final value achieved at the end of the search.

\subsection{Comparative Search Evaluation}
\label{sec:comparison}

We evaluate the effectiveness of \textbf{ELASTIC} by comparing it against three baseline NAS strategies: \textit{backbone-only search}~\cite{chen_detnas_2019}, \textit{head-only search}~\cite{wang_nas-fcos_2020}, and \textit{progressive NAS}~\cite{wang_nas-fcos_2020}. All methods are evaluated under a fixed budget using a consistent once-for-all supernet design.

\paragraph{Results on SVHN and PascalVOC subset.} Table~\ref{tab:main_results} summarizes our search results. On SVHN, ELASTIC achieves a final mAP of \textbf{80.09\%}, outperforming progressive NAS (79.62\%) while using only 12.5 GPU hours—a \textbf{2$\times$ speedup} compared to the 30.8 hours required by progressive search. On PascalVOC, the improvement is more substantial: ELASTIC reaches \textbf{30.83\% mAP}, outperforming progressive NAS (\textbf{21.74\%}) by \textbf{+9.09 percentage points} with similar compute (14.65 vs. 14.78 GPU hours).

\paragraph{Effect of Decoupling.} To isolate the impact of coordinated multi-module optimization, we compare ELASTIC against decoupled strategies that search only the backbone or the head. On SVHN, these methods achieve significantly lower performance: backbone-only search yields \textbf{75.53\% mAP} and head-only search reaches just \textbf{75.35\%}, falling short of ELASTIC's \textbf{80.09\% by 4.75 percentage points}. 

\paragraph{Convergence Behavior.} Figure~\ref{fig:history} shows that ELASTIC converges up to \textbf{2$\times$ faster} than progressive NAS. On SVHN, ELASTIC converges in just \textbf{15 iterations}, compared to \textbf{over 30} iterations for progressive NAS. In contrast, backbone-only and head-only searches stagnate early, \textbf{plateauing at 75.53\% and 75.35\%} mAP, respectively. A similar trend holds on PascalVOC, where ELASTIC rapidly reaches \textbf{30.83\% mAP}, outperforming progressive NAS (\textbf{21.74\%}) and single-module baselines (\textbf{22.02\% and 19.22\%}). This disparity is exacerbated by progressive NAS's rigid search pipeline, which fixes early module choices and prevents adaptation to downstream module configurations.

\paragraph{Comparison with MCUNET and TinyissimoYOLO.}
To assess ELASTIC’s suitability for microcontrollers, we extend our experiments using the MCUNET~\cite{lin_mcunet_2020} search space with the neck module included for fair comparison. Benchmarked on the full PascalVOC ($224 \times 224$) dataset with similar parameter budgets (Table~\ref{tab:mcunet}), our ELASTIC-derived detector achieves \textbf{72.3\%} mAP, surpassing MCUNET by \textbf{20.9\%}, TinyissimoYOLO’s~\cite{moosmann_tinyissimoyolo_2023} best model by \textbf{16.3\%}, and MCUNetV2-M4/H7~\cite{lin_mcunetv2_2024} by \textbf{7.7\%} and \textbf{4.0\%}, respectively, while using \textbf{50–75\% fewer MACs}. Unlike MCUNetV2~\cite{lin_mcunetv2_2024}, which requires external patch-based inference, ELASTIC supports native end-to-end inference, making it easier to deploy across diverse microcontrollers.

\begin{table*}[t]
    \centering
    \small
    \setlength{\tabcolsep}{6pt}
    \renewcommand{\arraystretch}{1.1}
    \caption{\textbf{Accuracy and efficiency across datasets and MCUs.} We report parameter count, detection accuracy, and per-inference efficiency for models on MAX78000/02 and STM32F746. Energy and power are omitted for STM32F746 due to setup limitations. Note we scale models deployed on the MAX78002, preserving key architectural choices, to deploy for comparison on the STM32F746.}
    \begin{tabular}{lccccccc}
        \toprule
        Model & Dataset & Device & Params & Energy ($\mu$J) & Latency (ms) & Power (mW) & mAP (\%) \\
        \midrule

        % ---------------- PascalVOC ----------------
        ai87-fpndetector~\cite{ai85tinierssd}  & PascalVOC   & MAX78002 & 2.18M & 62001 & 122.6 & 445.76 & 50.66 \\
        \textbf{ELASTIC (OURS)}                & PascalVOC   & MAX78002 & \textbf{1.32M} & \textbf{17581} & \textbf{51.1}  & \textbf{285.02} & \textbf{57.65} \\
        \addlinespace[2pt]
        ai87-fpndetector~\cite{ai85tinierssd}  & PascalVOC & STM32746 & 0.82M & N/A & 494.8 & N/A & 9.12 \\
        \textbf{ELASTIC (OURS)}                & PascalVOC & STM32746 & 0.89M & N/A & 490.6 & N/A & 19.0 \\
        \midrule

        % ---------------- TrashNet ----------------
        ai87-fpndetector~\cite{ai85tinierssd}  & TrashNet    & MAX78002 & 2.18M & 62001 & 122.6 & 445.76 & 83.1 \\
        \textbf{ELASTIC (OURS)}                & TrashNet    & MAX78002 & \textbf{1.32M} & \textbf{17581} & \textbf{51.1}  & \textbf{285.02} & \textbf{93.3} \\
        \addlinespace[2pt]
        ai87-fpndetector~\cite{ai85tinierssd}  & TrashNet & STM32746 & 0.82M & N/A & 494.8 & N/A & 54.03 \\
        \textbf{ELASTIC (OURS)}                & TrashNet & STM32746 & 0.89M & N/A & 490.6 & N/A & 73.9 \\
        \midrule

        % ---------------- VGGFace2 ----------------
        ai85net-tinierssd-face~\cite{ai85tinierssd} & VGGFace2 & MAX78000 & 0.28M & 1712 & \textbf{43.4} & 29.90 & 84.72 \\
        \textbf{ELASTIC (OURS)}                      & VGGFace2 & MAX78000 & \textbf{0.22M} & \textbf{1368} & 45.6 & \textbf{20.90} & \textbf{87.10} \\
        \addlinespace[2pt]
        ai85net-tinierssd-face~\cite{ai85tinierssd} & VGGFace2 & STM32746 & 0.277M & N/A & 968.2 & N/A & 80.2 \\
        \textbf{ELASTIC (OURS)}                      & VGGFace2 & STM32746 & \textbf{0.168M} & N/A & \textbf{699.1} & N/A & 82.1 \\
        \midrule

        % ---------------- SVHN ----------------
        ai85net-tinierssd~\cite{ai85tinierssd} & SVHN & MAX78000 & \textbf{0.19M} & 573 & 14.0 & 29.20 & 83.60 \\
        \textbf{ELASTIC (OURS)}                & SVHN & MAX78000 & 0.22M & \textbf{341} & \textbf{13.0} & \textbf{16.70} & \textbf{88.10} \\
        \addlinespace[2pt]
        ai85net-tinierssd~\cite{ai85tinierssd} & SVHN & STM32746 & 0.229M & N/A & 449.3 & N/A & 83.5 \\
        \textbf{ELASTIC (OURS)}                & SVHN & STM32746 & \textbf{0.235M} & N/A & 523.7 & N/A & \textbf{87.9} \\

        \bottomrule
    \end{tabular}
    \label{tab:max78000_results}
\end{table*}

\begin{figure}[t]
    \centering
    \includegraphics[width=0.9\linewidth]{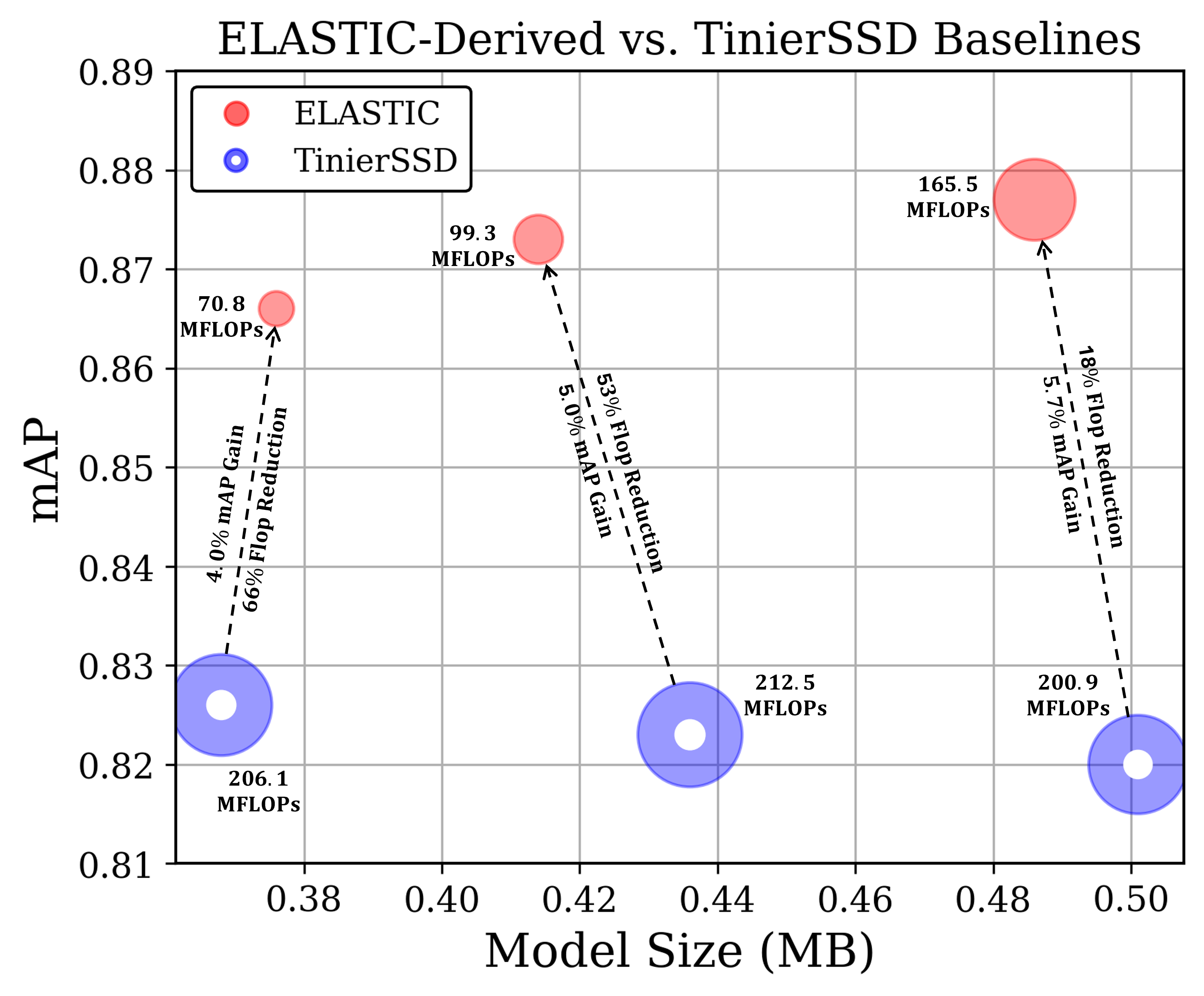}
    \caption{\textbf{Comparison of ELASTIC-derived models vs. scaled ai85net-tinierssd~\cite{ai85tinierssd} baselines.} Marker size reflects number of FLOPs, positively correlated with energy. ELASTIC-derived models can reduce FLOPs by up to 66\% with up to a 5.7\% accuracy gain.}
    \label{fig:sized}
\end{figure}

\subsection{Impact of Population Passthrough on Search Stability}
\label{sec:passthrough}

% Iterative NAS often suffers from performance drops when switching between modules due to population reinitialization. Without memory, each alternation incurs a “catch-up” phase, degrading search stability and slowing convergence. To address this, we introduce \textit{Population Passthrough}, a memory mechanism that retains top-performing architectures across module alternations.

% Figure~\ref{fig:passthrough} shows the effect of varying passthrough ratios on the PascalVOC dataset. Without passthrough, mAP repeatedly drops after each module switch and \textbf{saturates below 22\%}. In contrast, ratios of 50--80\% yield smoother trajectories and \textbf{reach over 30\% mAP within half the iterations}. The best results are obtained with a \textbf{60\% passthrough ratio}, which strikes an effective balance between \textit{inheritance} of strong candidates and \textit{exploration} of new architectures. Higher ratios risk overfitting to stale subnets, while lower ratios lead to instability.

Iterative NAS can suffer sharp drops at module boundaries because switching the active module effectively resets the evolutionary population. Without population memory, each alternation incurs a repeated ``catch-up'' phase to re-discover compatible high-quality candidates, harming stability and slowing convergence. We mitigate this with \textit{Population Passthrough}, which carries a fixed fraction of elite architectures into the next stage to preserve optimization momentum.

Figure~\ref{fig:passthrough} ablates the passthrough ratio on PascalVOC. With \textbf{0\%} passthrough, mAP collapses after switches and \textbf{saturates below 22\%}. Ratios of \textbf{50--80\%} stabilize the search and reach \textbf{$>30\%$} mAP in roughly half the iterations. \textbf{60\%} performs best, balancing exploitation (stability) and exploration (progress).

\subsection{Iterative Search Space Refinement}
\label{sec:refinement}

% To evaluate how ELASTIC improves the underlying search space over time, we conduct a series of random sampling experiments across both SVHN and PascalVOC. Specifically, we compare the distribution of model accuracies under three search conditions: (1) the original joint search space, (2) the progressive head-only search space, and (3) the refined head space produced by ELASTIC after iterative search.

% Figure~\ref{fig:sampling} shows the accuracy scatter plots and means of 100 randomly sampled architectures from each space. On SVHN, the joint space produces models with a \textbf{mean accuracy of 44.61\%}, while progressive head-only search \textbf{improves this to 67.38\%}. ELASTIC further enhances the head space, achieving a \textbf{mean accuracy of 72.25\% — a 4.87\% increase over progressive search}. A similar trend is observed on PascalVOC, where the mean accuracy \textbf{improves from 4.99\% (joint)} to\textbf{ 27.89\% (progressive)}, and then to \textbf{28.32\% with ELASTIC}.

% To further examine the progression of search space quality, we track mean and variance over time using random samples taken at key iterations of ELASTIC. As shown in Figure~\ref{fig:errorplot} and Table~\ref{tab:mean_std_variance}, the mean accuracy on SVHN \textbf{improves from 67.38\% at iteration 0 to 70.68\% at iteration 5}, while variance decreases, indicating a more stable and optimized search space. Similar patterns are observed on PascalVOC, where the \textbf{mean increases from 27.89\% to 28.32\%}, with \textbf{variance decreasing from $9.86 \times 10^{-5}$ to $6.60 \times 10^{-5}$}.

To evaluate whether ELASTIC improves the \emph{effective} search space, we randomly sample 100 architectures from three \emph{matched} spaces: (1) \emph{joint} backbone--neck/head, (2) \emph{progressive} head-only (fixed backbone), and (3) \emph{ELASTIC-refined} head-only (after re-optimization) in Figure~\ref{fig:sampling}. On SVHN, the mean mAP increases from \textbf{44.61\%} (joint) to \textbf{67.38\%} (progressive) and to \textbf{72.25\%} with ELASTIC (\textbf{+4.87\%} vs.\ progressive). On PascalVOC, the mean improves from \textbf{4.99\%} (joint) to \textbf{27.89\%} (progressive) and to \textbf{28.32\%} with ELASTIC.

We further track refinement over iterations by repeating the sampling at iterations \textbf{0, 3, 5} (Fig.~\ref{fig:errorplot}, Table~\ref{tab:mean_std_variance}), treating iteration \textbf{0} as the progressive baseline. On SVHN, the mean mAP rises from \textbf{67.38\%} (iter.\ 0) to \textbf{70.68\%} (iter.\ 5), with variance changing from \textbf{0.136} to \textbf{0.199}. On PascalVOC, the mean increases from \textbf{27.89\%} to \textbf{28.32\%}, while variance decreases from \textbf{$9.86\times10^{-5}$} to \textbf{$6.60\times10^{-5}$}. Overall, iterative revisiting yields a stronger induced space than progressive partitioning and improves stability on PascalVOC.

\subsection{Deployment on MCU Platforms}
\label{sec:deployment}

The MAX78000/MAX78002 MCUs integrate a CNN accelerator but impose tight constraints on kernel sizes, layer counts, and memory. Despite these limits, ELASTIC-derived models improve the efficiency--accuracy trade-off over TinierSSD/FPN baselines~\cite{ai85tinierssd} while remaining within on-chip budgets.

% We highlight these results in Table~\ref{tab:max78000_results}. On PascalVOC, energy decreases by \textbf{71.6\%}, latency by \textbf{2.40$\times$}, power by \textbf{36.1\%}, and mAP increases by \textbf{+6.99}. On TrashNet, ELASTIC follows the same pattern: energy decreases by \textbf{71.6\%}, latency by \textbf{2.40$\times$}, power by \textbf{36.1\%}, and mAP increases by \textbf{+10.2}. ELASTIC also generalizes across tasks: on VGGFace2, it reduces energy and power by \textbf{20.1\%} and \textbf{30.1\%}, respectively, while improving mAP by \textbf{+2.38}; on SVHN, it reduces energy and power by \textbf{40.5\%} and \textbf{42.8\%}, respectively, with a \textbf{+4.50} mAP gain and a slight latency improvement. We further evaluate cross-device portability on STM32F746. On VGGFace2, ELASTIC reduces latency by \textbf{27.8\%} and improves mAP by \textbf{+1.9}. On SVHN, ELASTIC improves mAP by \textbf{+4.4}.

We highlight these results in Table~\ref{tab:max78000_results}. On PascalVOC, energy decreases by \textbf{71.6\%}, latency by \textbf{2.40$\times$}, power by \textbf{36.1\%}, and mAP increases by \textbf{+6.99} on MAX78002. On TrashNet, ELASTIC follows the same pattern on MAX78002: energy decreases by \textbf{71.6\%}, latency by \textbf{2.40$\times$}, power by \textbf{36.1\%}, and mAP increases by \textbf{+10.2}. ELASTIC also generalizes across tasks: on VGGFace2, it reduces energy and power by \textbf{20.1\%} and \textbf{30.1\%}, respectively, while improving mAP by \textbf{+2.38}; on SVHN, it reduces energy and power by \textbf{40.5\%} and \textbf{42.8\%}, respectively, with a \textbf{+4.50} mAP gain and a slight latency improvement. We further evaluate cross-device portability on STM32F746. On PascalVOC, the scaled ELASTIC model reduces latency slightly from \textbf{494.8\,ms} to \textbf{490.6\,ms} while improving mAP from \textbf{9.12} to \textbf{19.0}. On TrashNet, the scaled ELASTIC model reduces latency slightly from \textbf{494.8\,ms} to \textbf{490.6\,ms} while improving mAP from \textbf{54.03} to \textbf{73.9}. On VGGFace2, ELASTIC reduces latency by \textbf{27.8\%} and improves mAP by \textbf{+1.9}. On SVHN, ELASTIC improves mAP by \textbf{+4.4}.

Since the two scaled STM32F746 models differ in their maximum deployable input resolution, we also compare them at both \textbf{$100\times100$} and \textbf{$128\times128$} in Table~\ref{tab:extra}. The scaled \texttt{ai87-fpndetector} baseline can be deployed only up to \textbf{$100\times100$}, while the scaled \textbf{ELASTIC} model remains deployable at \textbf{$128\times128$}; at \textbf{$128\times128$}, the baseline encounters out-of-memory due to activation SRAM limits on STM32F746. Under matched \textbf{$100\times100$} resolution, ELASTIC still outperforms the baseline on both datasets, improving mAP from \textbf{9.12} to \textbf{19.0} on PascalVOC and from \textbf{54.03} to \textbf{73.9} on TrashNet, while maintaining nearly identical latency (\textbf{490.6\,ms} vs.\ \textbf{494.8\,ms}). Moreover, ELASTIC supports the higher \textbf{$128\times128$} resolution and achieves \textbf{40.1} mAP on PascalVOC and \textbf{86.0} mAP on TrashNet, further highlighting that the searched architecture is not only more accurate but also more deployable under tight STM32F746 memory constraints. 

Figure~\ref{fig:sized} compares ELASTIC architectures to scaled TinierSSD baselines~\cite{ai85tinierssd} and shows ELASTIC attains higher accuracy at lower compute (\textbf{66\%} FLOPs reduction), underscoring the value of hardware-aware NAS for MCU deployment.

\begin{table*}[t]
    \centering
    \small
    \caption{\textbf{Accuracy comparison on STM32F746 at multiple input resolutions.} Note we scale these models down, preserving key architectural choices, to deploy on the STM32F746. The maximum input resolution ai87-fpndetector can deploy on is $(100\times100)$ while ELASTIC is $(128\times128)$.}
    \begin{tabular}{lcccccc}
        \toprule
        Model & Dataset & Device & Params & Resolution & Latency (ms) & mAP (\%) \\
        \midrule

        ai87-fpndetector~\cite{ai85tinierssd}  & PascalVOC & STM32746 & 0.82M & 128$\times$128 & N/A & OOM \\
        ai87-fpndetector~\cite{ai85tinierssd}  & PascalVOC & STM32746 & 0.82M & 100$\times$100 & 494.8 & 9.12 \\
        \textbf{ELASTIC (OURS)}                & PascalVOC & STM32746 & 0.89M & 128$\times$128 & 825.55 & 40.1 \\
        \textbf{ELASTIC (OURS)}                & PascalVOC & STM32746 & 0.89M & 100$\times$100 & 490.6 & 19.0 \\
        \midrule

        ai87-fpndetector~\cite{ai85tinierssd}  & TrashNet & STM32746 & 0.82M & 128$\times$128 & N/A & OOM \\
        ai87-fpndetector~\cite{ai85tinierssd}  & TrashNet & STM32746 & 0.82M & 100$\times$100 & 494.8 & 54.03 \\
        \textbf{ELASTIC (OURS)}                & TrashNet & STM32746 & 0.89M & 128$\times$128 & 825.55 & 86.0 \\
        \textbf{ELASTIC (OURS)}                & TrashNet & STM32746 & 0.89M & 100$\times$100 & 490.6 & 73.9 \\
        \bottomrule
    \end{tabular}
    \label{tab:extra}
\end{table*}

\begin{figure}[t]
    \centering
    \includegraphics[width=0.75\linewidth]{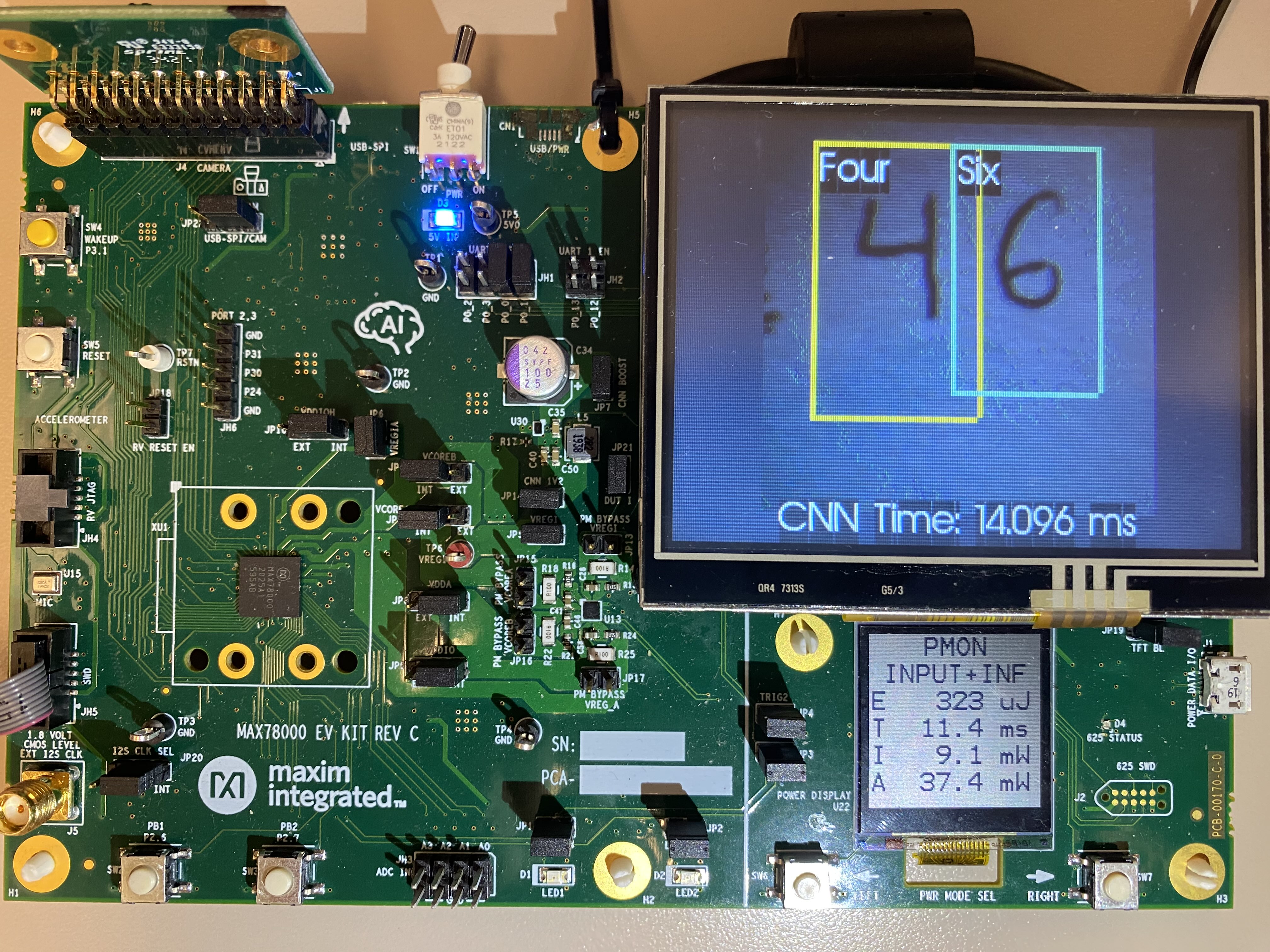}
    \caption{\textbf{ELASTIC deployment on MAX78000 using SVHN dataset.} ELASTIC produces compact, high-performing architectures suitable for ultra-low-power deployment.}
    \label{fig:max78000_device}
\end{figure}

\section{Conclusion}
\label{sec:conclusion}

We introduced \textbf{ELASTIC}, a unified framework for constrained, once-for-all NAS tailored to object detection on resource-limited platforms. By alternating backbone and head optimization, ELASTIC enables cross-module co-adaptation while significantly reducing search cost. A key component of our approach, \textit{Population Passthrough}, stabilizes module transitions and accelerates convergence, addressing a major limitation of existing progressive and decoupled NAS methods.

Through extensive experiments on multiple datasets and deployment on real microcontroller platforms, we demonstrated that ELASTIC discovers architectures that are more accurate and more efficient than manually designed and standard NAS baselines. These results highlight the effectiveness of iterative, hardware-aware NAS in advancing the accuracy–efficiency frontier for resource-constrained edge devices.

  % \paragraph{Limitations.} While ELASTIC reduces global NAS overhead and captures cross-module dependencies, its performance is sensitive to design choices such as search space, scheduling, alternation frequency, constraint assignment, and passthrough ratio. Our approach currently alternates only between backbone and head; extending to three modules (e.g., adding a neck) introduces coordination complexity. Furthermore, our evaluations are limited to small-to-medium datasets due to the constraints of TinyML deployment, which may not generalize directly to large-scale detection tasks.

  % \paragraph{Broader Impacts.} Our work enables efficient object detection on ultra-low-power devices by reducing both energy and latency, making it ideal for real-time, low-power applications. By enabling on-device inference, it enhances data privacy and eliminates reliance on cloud connectivity. These capabilities support sustainable, private, and scalable AI deployments in resource-constrained environments.
\balance
\bibliographystyle{IEEEtran}
\bibliography{IEEEtran}

% Biography text here without a photo.

% \begin{IEEEbiography}[{\includegraphics[width=1in,height=1.25in,clip,keepaspectratio]{fig1.png}}]{IEEE Publications Technology Team}
% In this paragraph you can place your educational, professional background and research and other interests.\end{IEEEbiography}

\end{document}